\documentclass[11pt]{article}
\usepackage{acl2002}
\usepackage{times}
\usepackage{latexsym}
\title{Evaluating the Effectiveness of Ensembles of Decision Trees\\
in Disambiguating Senseval Lexical Samples}
\author{Ted Pedersen\\
Department of Computer Science\\
University of Minnesota\\
Duluth, MN, 55812 USA\\
{\tt tpederse@d.umn.edu}
}

\date{}

\begin{document}                       
\maketitle
\begin{abstract}
This paper presents an evaluation of an ensemble--based system that  
participated in the English and Spanish lexical sample tasks of  
\textsc{Senseval-2}. The system combines decision trees  of unigrams,  
bigrams, and co--occurrences into a single classifier.
The analysis is extended to include the \textsc{Senseval-1} data. 
\end{abstract}

\section{Introduction}

There were eight Duluth systems that participated in the English and  
Spanish lexical sample tasks of \textsc{Senseval-2}. These systems were 
all based on the combination of lexical features with standard machine 
learning algorithms. The most accurate of these systems proved to be  
Duluth3 for English and Duluth8 for Spanish. These only differ with  
respect to minor language specific issues, so we refer to them  
generically as Duluth38, except when the language distinction is important. 

Duluth38 is an ensemble approach that assigns a sense to an instance of 
an ambiguous word by taking a vote among three bagged decision trees. Each 
tree is learned from a different view of the training examples associated  
with the target word.
Each view of the training examples is based on one of the following three  
types of lexical features: single words, two word sequences that occur  
anywhere within the context of the word being disambiguated, and two word  
sequences made up of this target word and another word within one or two  
positions. These features are referred to as unigrams, bigrams, and 
co--occurrences.

The focus of this paper is on determining if the member classifiers in  
the Duluth38 ensemble are complementary or redundant with 
each other and with other participating systems.  Two  classifiers are  
complementary if they disagree on a substantial number of disambiguation 
decisions and yet attain comparable levels of overall accuracy.  
Classifiers are redundant if they  arrive at the same disambiguation  
decisions for most  instances of the ambiguous word. There is little  
advantage in creating an ensemble of redundant classifiers, since they  
will make the same disambiguation decisions collectively as they would 
individually. An ensemble can only improve upon the accuracy of its member 
classifiers if they are complementary to  each other, and the errors of 
one classifier are offset by the correct  judgments of others. 

This paper continues with a description of the lexical features that make  
up the Duluth38 system, and then profiles the  \textsc{Senseval-1}  and 
\textsc{Senseval-2} lexical sample data that is used in this evaluation. 
There are two types of analysis presented. First, the accuracy of 
the member classifiers  in the Duluth38 ensemble are evaluated 
individually and in pairwise combinations. Second, the agreement between 
Duluth38 and the top two participating systems in \textsc{Senseval-1} and  
\textsc{Senseval-2} is compared. This paper concludes with a review
of the origins of our approach. Since the focus here is on 
analysis, implementation level details are not extensively discussed.  
Such descriptions can be found in  \cite{Pedersen01c} or 
\cite{Pedersen02a}.    

\section{Lexical Features}

Unigram features represent words that occur five or more times in the   
training examples associated with a given target word. A stop--list  
is used to eliminate high frequency function words as features.

For example, if the target word is {\it water} and the training example 
is {\it I water the flowering flowers}, the unigrams {\it water}, {\it  
flowering} and {\it flowers} are evaluated as possible unigram features.  
No stemming or other morphological processing is performed, so {\it  
flowering} and {\it flowers} are considered as distinct unigrams. {\it I}  
and {\it the} are not considered as possible features since they are 
included in the stop--list.   

Bigram features represent two word sequences that occur two or more times  
in the training examples associated with a target word,  and have a   
log--likelihood  value greater than or  equal to 6.635. This corresponds  
to a p--value of 0.01, which indicates that according to the  
log--likelihood ratio there is a 99\% probability that the words that 
make up this bigram are not independent.

If we are disambiguating {\it channel} and have the training example
{\it Go to the channel quickly}, then the three bigrams {\it Go to},
{\it the channel}, and {\it channel quickly} will be considered as  
possible features. {\it to the} is not included since both words are 
in the stop--list.             

Co--occurrence features are defined to be a pair of words that include the  
target word and another word within one or two positions. To be selected 
as a feature, a co--occurrence must occur two or more times in the lexical  
sample training data, and have a log--likelihood value of 2.706, which  
corresponds to a p--value of 0.10. A slightly higher p--value is used for  
the co--occurrence features, since the volume of data is much smaller 
than is available for the bigram features.

If we are disambiguating {\it art} and have the training example
{\it He and I like art of a certain period}, we evaluate {\it I art},
{\it like art}, {\it art of}, and {\it art a} as possible co--occurrence
features. 

All of these features are binary, and indicate if the designated unigram,  
bigram, or co--occurrence appears in the context with the ambiguous word. 
Once the features are identified from the training examples using the 
methods described above, the decision tree learner selects from among  
those features to determine which are most indicative of the sense of 
the ambiguous word. Decision tree learning is carried out with the Weka  
J48 algorithm \cite{weka}, which is a Java implementation of the  
classic C4.5 decision tree learner \cite{Quinlan86}.

\section{Experimental Data}

The English lexical sample for \textsc{Senseval-1} is made  
up of 35 words, six of which are used in multiple parts of speech. The  
training examples have been manually annotated based on the HECTOR
sense inventory. There are 12,465 training examples, and 7,448 test  
instances. This corresponds to what is known as the {\it trainable}  
lexical sample in the \textsc{Senseval-1} official results. 

The English lexical sample for \textsc{Senseval-2} consists of 73 word
types, each of which is associated with a single part of
speech. There are 8,611 sense--tagged  examples provided for training,
where each instance has been manually  assigned a WordNet sense. The
evaluation data for the English  lexical sample consists of 4,328 held
out test instances.   

The Spanish lexical sample for \textsc{Senseval-2} consists of 39 word  
types. There are 4,480 training examples that have been manually tagged  
with senses from Euro-WordNet. The evaluation data consists of 2,225 test  
instances.  

\section{System Results}

This section (and Table 1) summarizes the performance of the top two  
participating systems in \textsc{Senseval-1} and \textsc{Senseval-2}, as 
well as the Duluth3 and Duluth8 systems. Also included are baseline 
results for a decision stump and a majority classifier. A decision stump  
is simply a one node decision tree based on a co--occurrence feature,   
while the majority classifier assigns the most frequent sense in the  
training data to every occurrence of that word in the test data. 

Results are expressed using accuracy, which is computed by dividing the  
total number of correctly disambiguated test instances by the total number  
of test instances. Official results from \textsc{Senseval} are reported  
using precision and recall, so these are converted to accuracy to provide  
a consistent point of comparison. We utilize fine grained scoring, where a  
word is considered correctly disambiguated only if it is assigned exactly  
the sense indicated in the manually created gold standard. 

In the English lexical sample task of \textsc{Senseval-1} the two most
accurate systems overall were hopkins-revised (77.1\%) and
ets-pu-revised (75.6\%). The Duluth systems did not participate in this 
exercise, but have been evaluated using the same data after the fact.
The Duluth3 system reaches accuracy of 70.3\%. The simple majority  
classifier attains accuracy of 56.4\%.  

In the English lexical sample task of \textsc{Senseval-2} the two most 
accurate systems were JHU(R) (64.2\%) and SMUls (63.8\%).  Duluth3 attains  
an accuracy of 57.3\%, while a simple majority classifier attains accuracy  
of 47.4\%.  

In the Spanish lexical sample task of \textsc{Senseval-2} the two most
accurate systems were JHU(R) (68.1\%) and stanford-cs224n (66.9\%).   
Duluth8 has accuracy of 61.2\%, while a simple majority  
classifier attains accuracy of 47.4\%.   

The top two systems from the first and second \textsc{Senseval}
exercises represent a wide range of strategies that we can only hint
at here. The SMUls English lexical sample system is perhaps the most
distinctive in that it incorporates information from WordNet, the
source of the sense distinctions in \textsc{Senseval-2}. The
hopkins-revised, JHU(R), and stanford-cs224n systems use supervised
algorithms that learn classifiers from a rich combination of syntactic
and lexical features. The ets-pu-revised system may be the closest in
spirit to our own, since it creates an ensemble of two Naive Bayesian
classifiers, where one is based on topical context and the other on 
local context. 

More detailed description of the \textsc{Senseval-1} and  
\textsc{Senseval-2} systems and lexical samples can be found in  
\cite{KilgarriffP00} and  \cite{EdmondsC01}, respectively. 

\begin{table}
\caption{Accuracy in Lexical Sample Tasks} 
\begin{tabular}{lcr}
\hline
\hline
system & accuracy & correct \\
\hline
 & & \\
\multicolumn{3}{c}{English \textsc{Senseval-1}} \\
 & & \\
hopkins-revised & 77.1\% & 5,742.4 \\ 
ets-pu-revised & 75.6\% & 5,630.7 \\
UC  & 71.3\%   &  5,312.8 \\
UBC & 70.3\%   &  5,233.9 \\
BC  & 70.1\%   &  5,221.7 \\
UB  & 69.5\%   &  5,176.0 \\
C   & 69.0\%   &  5,141.8 \\
B   & 68.1\%   &  5,074.7 \\
U   & 63.6\%   &  4,733.7 \\
stump  & 60.7\% & 4,521.0 \\
majority & 56.4\% & 4,200.0 \\
\hline
 & & \\
\multicolumn{3}{c}{English \textsc{Senseval-2}} \\
 & & \\
JHU(R) & 64.2\% &  2,778.6 \\
SMUls & 63.8\% &  2,761.3 \\
UBC & 57.3\% &  2,480.7   \\
UC &   57.2\% &  2,477.5 \\
BC &  56.7\% &  2,452.0 \\
C   & 56.0\%   &  2,423.7 \\
UB &   55.6\% &  2,406.0 \\
B &   54.4\% &  2,352.9 \\
U   & 51.7\% &  2,238.2 \\
stump  & 50.0\% &  2,165.8 \\
majority & 47.4\% & 2,053.3 \\
\hline
 & & \\
\multicolumn{3}{c}{Spanish \textsc{Senseval-2}} \\
 & & \\
JHU(R) & 68.1\% & 1,515.2 \\
stanford-cs224n & 66.9\% & 1,488.5 \\ 
UBC & 61.2\% & 1,361.3 \\
BC & 60.1\% & 1,337.0 \\
UC & 59.4\% & 1,321.9 \\
UB & 59.0\% & 1,312.5 \\
B & 58.6\% & 1,303.7 \\
C   & 58.6\% & 1,304.2 \\
stump &  52.6\% & 1,171.0 \\
U   &  51.5\% & 1,146.0 \\
majority &  47.4\% & 1,053.7 \\
\hline
\end{tabular}
\label{results-table}
\end{table}

\section{Decomposition of Ensembles}

The three bagged decision trees that make up Duluth38 are evaluated
both individually and as pairwise ensembles. In Table  
\ref{results-table} and subsequent discussion, we refer to the  
individual bagged decision trees based on unigrams, bigrams and  
co--occurrences as U, B, and C, respectively. We designate ensembles that  
consist of two or three bagged decision trees by using the relevant  
combinations of letters. For example, UBC refers to a three member  
ensemble consisting of unigram (U), bigram (B), and co--occurrence (C)  
decision trees, while BC refers to a two member ensemble of bigram (B) and  
co-occurrence (C) decision trees. Note of course that UBC is synonymous 
with Duluth38.  

Table \ref{results-table} shows that Duluth38 (UBC) achieves 
accuracy significantly better than the lower bounds represented by the  
majority classifier and the decision stump, and comes within seven  
percentage points of the most accurate systems in each of the three  
lexical sample tasks. However, UBC does not significantly improve upon
all of its member classifiers, suggesting that the ensemble is made up 
of redundant rather than complementary classifiers. 

In general the accuracies of the bigram (B) and co--occurrence (C)  
decision  trees are never significantly different than the accuracy 
attained by the ensembles of which they are members (UB, BC, UC,  
and UBC), nor are they significantly different from each other. This is  
an intriguing result, since the co--occurrences represent a much smaller 
feature set than bigrams, which are in turn much smaller than the unigram  
feature set.  Thus, the smallest of our feature sets is the  
most effective. This may be due to the fact that small feature sets are  
least likely to suffer from  fragmentation during decision tree learning. 

Of the three individual bagged decision trees U, B, and C, the unigram  
tree (U) is significantly less accurate for all three lexical samples. It  
is only slightly more accurate than the decision stump for both English  
lexical samples, and is less accurate than the decision stump in the 
Spanish task. 

The relatively poor performance of unigrams can be accounted for by the  
large number of possible features. Unigram features consist of all words 
not in the stop--list that occur five or more times in the training  
examples for a word. The decision tree learner must search through a very 
large feature space, and under such circumstances may fall victim to 
fragmentation. 

Despite these results, we are not prepared to dismiss the use of  
ensembles or unigram decision trees. An ensemble of unigram and 
co--occurrence decision  trees (UC)  results in greater accuracy than any 
other lexical decision tree for the English \textsc{Senseval-1} lexical 
sample, and is essentially tied with the most accurate of these approaches
(UBC) in the English \textsc{Senseval-2} lexical sample. In principle 
unigrams and co--occurrence features are complementary, since unigrams 
represent topical context, and co--occurrences represent local context. 
This follows the line of reasoning developed by \cite{LeacockCM98} in 
formulating their ensemble of Naive Bayesian classifiers for word sense 
disambiguation. 

Adding the bigram decision tree (B) to the ensemble of the unigram and  
co--occurrence decision trees (UC) to create UBC does not result in  
significant improvements in accuracy for the any of the lexical samples. 
This reflects the fact that the bigram and co--occurrence feature sets
can be  redundant. Bigrams are two
word sequences that occur anywhere within the context of the ambiguous  
word, while co--occurrences are bigrams that include the target word 
and a word one or two positions away. Thus, any consecutive two word 
sequence that includes the word to be disambiguated and has a  
log--likelihood ratio greater than the specified threshold will be  
considered both a bigram and a co--occurrence. 

Despite the partial overlap between bigrams and co--occurrences, we
believe that retaining them as separate feature sets is a reasonable idea. 
We have observed that an ensemble of multiple decision  
trees where each is learned from a representation of the training examples  
that has a small number of features is more accurate than a single  
decision tree that is learned from one large representation of the  
training examples. For example, we mixed the bigram and co--occurrence  
features into a single feature set, and then learned a single bagged  
decision tree from this representation of the training examples. We  
observed drops in accuracy in both the Spanish and English  
\textsc{Senseval-2} lexical sample tasks. For Spanish it falls from 59.4\%  
to 58.2\%, and for English it drops from 57.2\% to 54.9\%. Interestingly  
enough, this mixed feature set of bigrams and co--occurrences results in a  
slight increase over an ensemble of the two in the \textsc{Senseval-1}  
data, rising from 71.3\% to 71.5\%.      

\section{Agreement Among Systems}

The results in Table \ref{results-table} show that UBC and 
its member classifiers perform at levels of accuracy significantly higher 
than the majority classifier and decision stumps, and approach the level 
of some of the more accurate systems. This poses an intriguing  
possibility. If UBC is making  complementary errors to those other  
systems, then it might be possible to combine these systems to achieve an   
even higher level of accuracy. The alternative is that the decision trees   
based on lexical features are largely redundant with these other systems,  
and that there is a hard core of test instances that are resistant to   
disambiguation by any of these systems.  

We performed a series of pairwise comparisons to establish the degree to  
which these systems agree. We included the two most accurate participating  
systems from each of the three lexical sample tasks, along with UBC, a  
decision stump, and a majority classifier. 

\begin{table}
\caption{System Pairwise Agreement}
\begin{tabular}{lrrr}
\hline
\hline
\multicolumn{1} {l} {system pair} &
\multicolumn{1} {c} {both} & 
\multicolumn{1} {c} {one} & 
\multicolumn{1} {c} {zero} \\
\hline 
\multicolumn{4}{c} {}\\
\multicolumn{4} {c} {English \textsc{Senseval-1}} \\ \\
hopkins ets-pu & 67.8\% & 17.1\% & 12.1\% \\
 & 5,045 & 1,274 & 1,126 \\ \\
UBC hopkins & 64.8\% & 18.3\% & 17.0\%  \\
 & 4,821 & 1,361 & 1,263 \\ \\
UBC ets-pu & 64.4\% & 17.4\% & 18.2\% \\
 & 4,795 & 1,295 & 1,355 \\ \\
stump majority & 53.4\% & 13.7\% & 32.9\% \\
 & 3,974 & 1,022 & 2,448 \\  
\hline
\multicolumn{4}{c} {}\\
\multicolumn{4} {c} {English \textsc{Senseval-2}} \\ \\
JHU(R) SMUls & 50.4\% & 27.3\% & 22.3\% \\
 & 2,180 & 1,183 & 965 \\ \\
UBC JHU(R) & 49.2\% & 24.1\% & 26.8\% \\
 & 2,127 & 1,043 & 1,158 \\ \\
UBC SMUls & 47.2\% & 27.5\% & 25.2\% \\
 & 2,044 & 1,192 & 1,092 \\ \\
stump majority & 45.2\% & 11.8\% & 43.0\% \\
 & 1,955 & 511 & 1,862 \\ 
\hline
\multicolumn{4}{c} {}\\
\multicolumn{4} {c} {Spanish \textsc{Senseval-2}} \\ \\
JHU(R) cs224n & 52.9\% & 29.3\% & 17.8\% \\
 & 1,177 & 651 & 397 \\ \\
UBC cs224n & 52.8\% & 23.2\% & 24.0\% \\
 & 1,175 & 517 & 533 \\ \\
UBC JHU(R) & 48.3\% & 33.5\% & 18.2\% \\
 & 1,074 & 746 & 405 \\ \\
stump majority & 45.4\% & 20.4\% & 34.2\% \\ 
 & 1,011 & 453 & 761 \\
\hline
\end{tabular}
\label{table-agree}
\end{table}

In Table ~\ref{table-agree} the column labeled ``both'' shows the 
percentage and count of test instances where both systems are correct, 
the column labeled ``one'' shows the percentage and count where only one  
of the two systems is correct, and the column labeled ``none'' shows how  
many test instances were not correctly disambiguated by either system. 
We note that in the pairwise comparisons there is a high level of  
agreement for the instances that both systems were able to disambiguate,   
regardless of the systems involved. For example, in the   
\textsc{Senseval-1} results the three pairwise comparisons among UBC,  
hopkins-revised, and ets-pu-revised all show that approximately 65\% of  
the test instances are correctly disambiguated by both systems. The same  
is true for the English and Spanish lexical sample tasks in  
\textsc{Senseval-2}, where each pairwise comparison results in agreement  
in approximately half the test instances. 

Next we extend this study of agreement to a three--way comparison
between UBC, hopkins-revised, and ets-pu-revised for the
\textsc{Senseval-1} lexical sample.  There are 4,507 test instances
where all three systems agree (60.5\%), and 973 test instances
(13.1\%) that  none of the three is able to get correct. These are
remarkably similar values to the pair--wise comparisons, suggesting
that there is a fairly consistent number of test instances that all
three systems handle in the same way. When making a five--way
comparison that includes these three systems and the decision stump
and the majority classifier, the number of test instances that no
system can disambiguate correctly drops to 888, or 11.93\%.  This is
interesting in that it shows there are nearly 100 test instances that
are only disambiguated correctly by the decision stump or the majority
classifier, and not by any of the other three systems. This suggests
that very simple classifiers are able to resolve some test instances
that more complex techniques miss.   

The agreement when making a three way comparison between UBC, JHU(R),
and SMUls in the English \textsc{Senseval-2} lexical sample drops
somewhat from the pair--wise levels. There are 1,791 test instances
that all three systems disambiguate correctly (41.4\%) and 828
instances that none of these systems get correct (19.1\%). When making
a five way comparison between these three systems, the decision stump
and the majority classifier, there are 755 test instances (17.4\%)
that no system can resolve. This shows that these three systems are
performing somewhat differently, and do not agree as much as the
\textsc{Senseval-1} systems. 

The agreement when making a three way comparison between UBC, JHU(R),
and cs224n in the Spanish lexical sample task of \textsc{Senseval-2}
remains fairly consistent with the pairwise comparisons. There are 960
test instances that all three systems get correct (43.2\%), and 308
test instances where all three systems failed (13.8\%). When making a
five way comparison between these three systems and the decision stump
and the majority classifier, there were 237 test instances (10.7\%)
where no systems was able to resolve the sense. Here again we see
three systems that are handling quite a few test instances in the same
way. 

Finally, the number of cases where neither the decision stump nor the
majority classifier is correct varies from 33\% to 43\% across the
three lexical samples.  This suggests that the optimal combination of
a majority classifier and decision stump could attain overall accuracy
between 57\% and 66\%, which is comparable with some of the better
results for these lexical samples. Of course, how to achieve such an
optimal combination is an open question. This is still an interesting
point, since it suggests that there is a relatively large number of
test instances that require fairly minimal information to disambiguate
successfully.   

\section{Duluth38 Background}

The origins of Duluth38 can be found in an ensemble approach based on   
multiple Naive Bayesian classifiers that perform disambiguation via  
a majority vote \cite{Pedersen00b}. Each member of the ensemble is based
on unigram features  that occur in varying sized windows of context to
the left and right of the ambiguous word. The sizes of these windows
are 0, 1, 2, 3, 4, 5, 10, 25, and 50 words to the left and to the
right, essentially forming bags of words to the left and right. The
accuracy of this ensemble disambiguating the nouns {\it interest}
(89\%) and {\it line} (88\%) is as high as any previously published
results. However, each ensemble consists of 81 Naive Bayesian
classifiers, making it difficult to determine which features and 
classifiers were contributing most significantly to disambiguation. 

The frustration with models that lack an intuitive interpretation led 
to the development of decision trees based on bigram features   
\cite{Pedersen01b}. This is quite similar to the bagged decision  
trees of bigrams (B) presented here, except that the earlier work learns 
a single decision tree where training examples are represented by the top  
100 ranked bigrams, according to the log--likelihood ratio. This 
earlier approach was evaluated on the \textsc{Senseval-1} data and 
achieved an overall accuracy of 64\%, whereas the bagged decision tree 
presented here achieves an accuracy of 68\% on that data. 

Our interest in co--occurrence features is inspired by     
\cite{ChouekaL85}, who showed that humans determine the
meaning of ambiguous words largely based on words that occur within
one or two  positions to the left and right.  Co--occurrence features,
generically defined as bigrams where one of the words is the target
word and the other occurs within a few positions, have been widely used in  
computational approaches to word sense disambiguation. When the impact 
of mixed feature sets on disambiguation is analyzed, co--occurrences 
usually prove to contribute significantly to overall accuracy. This is 
certainly our experience, where the co--occurrence decision tree (C) is  
the most accurate of the individual lexical decision trees. Likewise, 
\cite{NgL96} report  overall accuracy for the noun {\it interest} of 87\%,  
and find that that when their feature set only consists of co--occurrence  
features the accuracy only drops to 80\%.       

Our interest in bigrams was indirectly motivated by \cite{LeacockCM98}, 
who describe an ensemble approach made up of local context and topical 
context. They suggest that topical context can be represented by words 
that occur anywhere in a window of context, while local contextual  
features are words that occur within close proximity to the target word.  
They show that in disambiguating the adjective {\it hard} and the verb  
{\it serve} that the local context is most important, while for 
the noun {\it line} the topical context is most important. We believe 
that statistically significant bigrams that occur anywhere in the window 
of context can serve the same role, in that such a two word sequence 
is likely to carry heavy semantic (topical) or syntactic (local)  
weight.

\section{Conclusion}

This paper analyzes the performance of the Duluth3 and Duluth8 systems 
that participated in the English and Spanish lexical sample tasks in 
\textsc{Senseval-2}. We find that an ensemble offers very limited  
improvement over individual decision trees based on lexical features.  
Co--occurrence decision trees are more accurate than bigram or unigram  
decision trees, and are nearly as accurate as the full ensemble. This is  
an encouraging  result, since the number of co--occurrence features is  
relatively small and easy to learn from compared to the number of bigram  
or unigram features. 

\section{Acknowledgments}

This work has been partially supported by a National Science Foundation
Faculty Early CAREER Development award (\#0092784). 
                                   
The Duluth38 system (and all other Duluth systems that participated in 
\textsc{Senseval-2}) can be downloaded from the author's web site: 
http://www.d.umn.edu/\~{}tpederse/code.html. 


\end{document}